\documentclass{article}

\usepackage{microtype}
\usepackage{graphicx}
\usepackage{subcaption}
\usepackage{booktabs} 
\usepackage{multirow}
\usepackage{hyperref}
\usepackage{microtype}
\usepackage{array}
\usepackage{bm}

\usepackage[preprint]{icml2026}

\usepackage{amsmath}
\usepackage{amssymb}
\usepackage{mathtools}
\usepackage{amsthm}
\usepackage{enumitem}

\usepackage[capitalize,noabbrev]{cleveref}

\theoremstyle{plain}

\theoremstyle{definition}

\theoremstyle{remark}

\usepackage[textsize=tiny]{todonotes}

\usepackage{graphicx} 
\usepackage{amssymb}
\begin{document}
\twocolumn[
  \icmltitle{Meta-Aligner: Bidirectional Preference-Policy Optimization for Multi-Objective LLMs Alignment}



  \icmlsetsymbol{equal}{*}

  \begin{icmlauthorlist}
    \icmlauthor{Wenzhe Xu}{equal}
    \icmlauthor{Biao Liu}{equal}
    \icmlauthor{Yiyang Sun}{}
    \icmlauthor{Xin Geng}{}
    \icmlauthor{Ning Xu}{}
  \end{icmlauthorlist}


  \icmlcorrespondingauthor{Ning Xu}{xning@seu.edu.cn}

  \icmlkeywords{Machine Learning, Large Language Models}

  \vskip 0.3in
]
\printAffiliationsAndNotice{\icmlEqualContribution}

\newcommand{\ours}{\textsc{Meal}}
\begin{abstract}
Multi-Objective Alignment aims to align Large Language Models (LLMs) with diverse and often conflicting human values by optimizing multiple objectives simultaneously. Existing methods predominantly rely on static preference weight construction strategies. However, rigidly aligning to fixed targets discards valuable intermediate information, as training responses inherently embody valid preference trade-offs even when deviating from the target. To address this limitation, we propose \ours, i.e., MEta ALigner, a bi-level meta-learning framework enabling bidirectional optimization between preferences and policy responses, generating instructive dynamic preferences for steadier training. Specifically, we introduce a preference-weight-net as a meta-learner to generate adaptive preference weights based on input prompts and update the preference weights as learnable parameters, while the LLM policy acts as a base-learner optimizing response generation conditioned on these preferences with rejection sampling strategy. Extensive empirical results demonstrate that our method achieves superior performance on several multi-objective benchmarks, validating the effectiveness of the dynamic bidirectional preference-policy optimization framework.
\end{abstract}
\section{Introduction}

Large Language Models (LLMs) have demonstrated remarkable capabilities across a wide range of natural language processing tasks \cite{liang2024controllable,wang2023enabling}, serving as powerful tools for reasoning, coding, and creative generation \cite{xu2025towards,jiang2024survey}. However, ensuring that LLMs align with human values and intentions, such as helpfulness, honesty, and harmlessness, remains a fundamental challenge in the development of responsible AI systems. The primary  paradigm for addressing this challenge is Reinforcement Learning from Human Feedback (RLHF) \cite{christiano2017deep,stiennon2020learning, ouyang2022training}. In the standard RLHF pipeline, a reward model is first trained to approximate human preferences, and the LLM is subsequently fine-tuned to maximize the expected reward using algorithms like PPO \cite{schulman2017proximal}.

This kind of approach oversimplifies the complexity of real-world applications, where LLMs are unavoidably required to satisfy diverse and often conflicting values simultaneously, such as balancing helpfulness with harmlessness, or navigating varying standards of verbosity and creativity. 
Recently, there has been growing interest in Multi-Objective Alignment, which aims to equip models with the capability to navigate trade-offs between competing objectives and adapt to diverse user preference profiles \cite{li2020deep,rame2023rewarded,yang2024rewards}. 
Early attempts relied on scalarization techniques, which aggregate multiple reward models using fixed weights to derive a single optimization target~\cite{li2020deep}. However, these static weighting schemes are inflexible and fail to accommodate the diversity of user preferences in real-world scenarios.
To mitigate this, contemporary studies have shifted towards test-time adaptation. Prominent approaches encompass parameter interpolation \cite{rame2023rewarded, shi2024decoding}, prompt-conditioned generation \cite{yang2024rewards, wang2024arithmetic}, and preference-aware reward modeling \cite{lin2025parm}.

However, existing multi-objective alignment methods predominantly rely on static preference weight construction strategies to guide model optimization. During training, generated responses may deviate from specified targets, yet they inherently embody valid preference trade-offs. Rigidly forcing alignment to static targets—especially in SFT-based methods—discards this valuable intermediate information. Therefore, rigid unidirectional optimization fails to capture the rich mapping between the policy's evolving responses and their underlying preferences. To address the limitation, we consider a more flexible bidirectional optimization paradigm, where the preference weights and the policy responses can dynamically adapt to each other during training, allowing the model to iteratively refine its preference understanding and improve the alignment performance.

Motivated by the above consideration, we propose \ours, i.e., MEta ALigner, a bi-level optimization framework that jointly trains a preference-weight-net and the LLM policy to achieve dynamic preference-policy alignment. The preference-weight-net serves as a meta-learner that generates adaptive preference weights based on the input prompt, while the LLM policy acts as a base-learner that optimizes its response generation conditioned on these preferences. Unlike prior methods that enforce a static mapping from fixed preference weights to responses, our framework establishes a dynamic feedback loop: the preference-weight-net learns to adjust its weight generation strategy based on the policy's evolving performance, while the policy continuously refines its response generation in light of the updated preferences. This bidirectional interaction enables the model to capture intermediate preference states, providing steadier instructive guidance during training
Our main contributions are summarized as follows:
\begin{enumerate}[leftmargin=*,itemsep=0pt]
    \item We propose \ours, a bi-level meta-learning framework that enables bidirectional optimization between preferences and policy responses, mitigating the loss of valuable intermediate preference information inherent in static weighting methods. 
    \item We develop an outer-loop adaptation strategy that updates the preference-weight-net based on the policy's generated responses, continuously adapting to the evolving state of the policy during training.
    \item We incorporate the preference-weight-net within an online learning paradigm, where the model generates responses conditioned on dynamically produced preferences and progressively improves via high-scoring sample selection.
\end{enumerate}
Extensive empirical results show that our method achieves better performance on complex multi-objective benchmarks, validating the effectiveness of the dynamic bidirectional preference-policy optimization framework.

\section{Related Work}

Aligning Large Language Models (LLMs) with human values is a fundamental prerequisite for building safe and reliable AI systems \cite{achiam2023gpt, chen2025reasoning}. 
The dominant paradigm, Reinforcement Learning from Human Feedback (RLHF) \cite{christiano2017deep, stiennon2020learning, ouyang2022training}, typically involves training a reward model to approximate human preferences and subsequently optimizing the policy via algorithms like PPO \cite{schulman2017proximal}. Despite its success, standard RLHF is often computationally expensive and unstable \cite{dong2023raft, yuan2023rrhf}. To mitigate these issues, recent advancements have shifted towards Direct Preference Optimization (DPO) \cite{rafailov2023direct}, which optimizes the policy directly on preference data without an explicit reward model. 
Building on this, researchers have developed more efficient variants to further enhance training stability and performance, such as SimPO \cite{meng2024simpo}, KTO \cite{ethayarajh2024model}, ORPO \cite{hong2024orpo}, and IPO \cite{garg2025ipo}.

While the aforementioned approaches have significantly improved alignment efficiency, they predominantly model human preference as a single scalar reward. 
This simplification overlooks the complexity of real-world applications, where LLMs must simultaneously satisfy diverse and often conflicting objectives, such as balancing helpfulness with harmlessness or navigating varying stylistic constraints \cite{liang2024controllable}. Consequently, a single optimization target is insufficient to capture the Pareto frontier of human values. Early attempts to address this challenge relied on scalarization \cite{li2020deep}, which aggregates multiple reward models using fixed weights.

To overcome the limitations of fixed scalarization, recent literature has explored more dynamic multi-objective alignment strategies. Extending the success of direct preference learning, methods like MODPO \cite{zhou2024beyond} and CPO \cite{guo2024controllable} adapt DPO-style objectives to multi-objective scenarios, avoiding the instability of RL-based optimization. Another growing line of research focuses on test-time adaptation to accommodate diverse user preferences. Notable strategies include interpolating parameters of expert models, such as Rewarded Soups \cite{rame2023rewarded} and MOD \cite{shi2024decoding}, or controlling generation via prompt conditioning and rejection sampling, as seen in Rewards-in-Context (RIC) \cite{yang2024rewards} and DPA \cite{wang2024arithmetic}. 
More recently, PARM \cite{lin2025parm} proposed a preference-aware autoregressive reward model to guide generation dynamically. 

\section{Preliminary}
We begin by formally defining the notation for the language model alignment task. Let $\mathcal{V}$ denote the vocabulary of the language model. The generation process is modeled by a policy $\pi: \mathcal{X} \rightarrow \mathcal{Y}$, which maps an input query $x \in \mathcal{X}$ to a corresponding response $y \in \mathcal{Y}$. Formally, the input query and the generated response are represented as sequences of tokens $\bm{x} = [x^1, x^2, \dots, x^m]$ and $\bm{y} = [y^1, y^2, \dots, y^n]$. Here, the input and output spaces are defined as $\mathcal{X} = \mathcal{V}^m$ and $\mathcal{Y} = \mathcal{V}^n$, where $m$ and $n$ denote the sequence lengths. The primary objective of alignment is to ensure that the policy $\pi$ generates responses $\bm{y}$ that are consistent with human values and preferences for a given context $\bm{x}$.

\begin{figure*}[t!]
\centering %
\vskip 0.2in
\centering
\includegraphics[width=\textwidth]{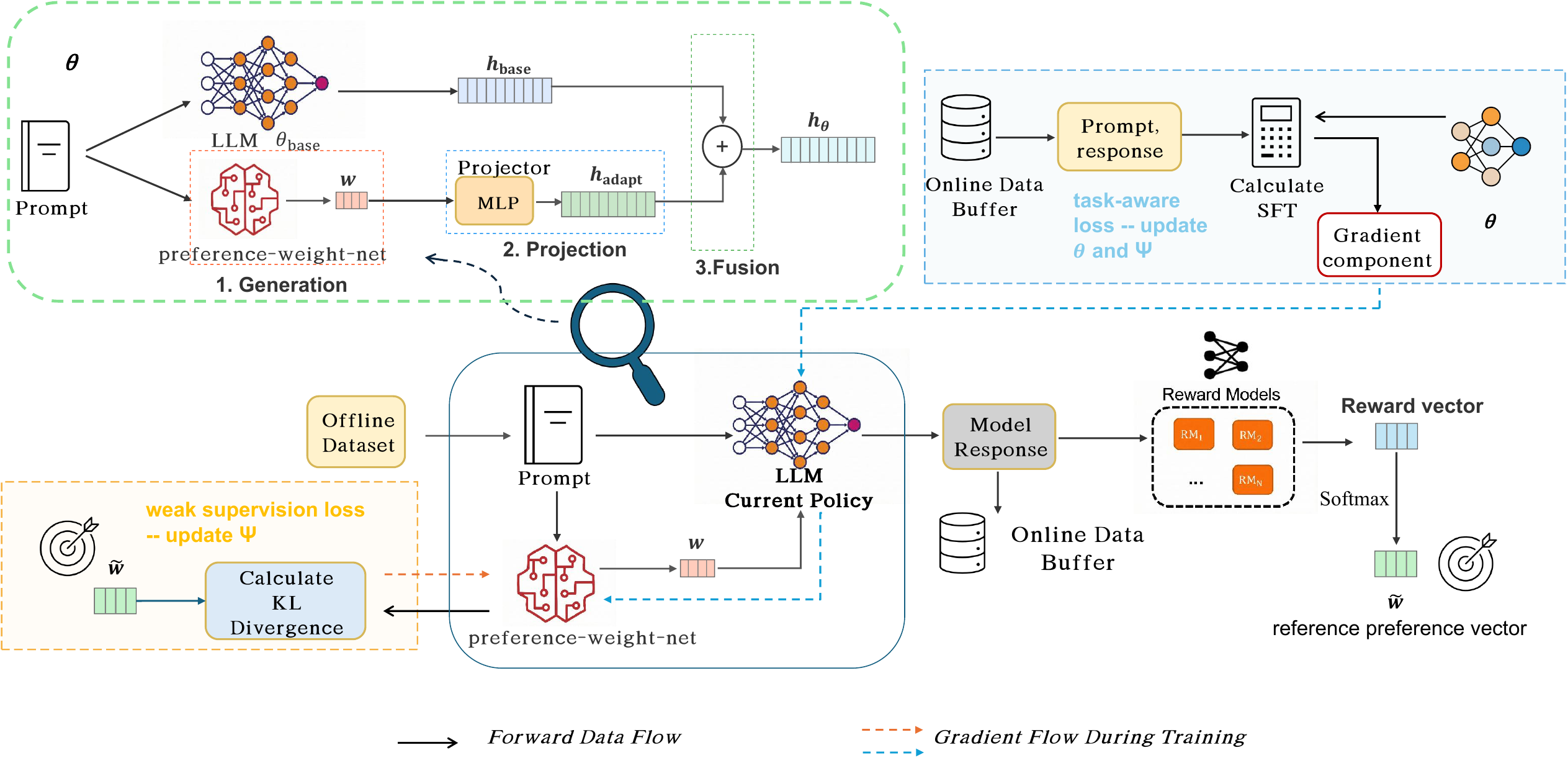}
\caption{Illustration of \ours}
\label{fig:method}
\end{figure*}

In real-world applications, an LLM often needs to satisfy multiple objectives (e.g., helpfulness, harmlessness, humorousness), which are even potentially conflicting. We assume there are $K$ distinct objectives, each associated with a reward model $r_k: \mathcal{X} \times \mathcal{Y} \rightarrow \mathbb{R}$.
For a specific input $\bm x$ and response $\bm y$, the multi-objective evaluation is often represented as a vector of reward scores $r(\bm x, \bm y) = [r_1(\bm x, \bm y), \dots, r_K(\bm x, \bm y)]$.
To balance these objectives, a preference weight vector $\bm{w} = [w_1, \dots, w_K] \in \Delta^{K-1}$ is introduced, where $\sum_{k=1}^K w_k = 1$, $w_k \ge 0$ ,and $w_k$ indicates the relative importance of the $k$-th objective.
The scalar reward for a specific preference configuration is defined as the weighted sum:
\begin{equation}
R(\bm x, \bm y, \bm w) = \sum_{k=1}^K w_k r_k(\bm x, \bm y) = \bm{w}^\top r(\bm x, \bm y)  
\end{equation}

\section{Method}
We propose \ours, a bi-level meta-learning framework for multi-objective alignment. 
As illustrated in \cref{fig:method}, the framework consists of two learnable components: the LLM policy $\pi_{{\theta}}$ and the preference-weight-net $f_{\psi}$. Here, we formally define the parameters of the policy as $\theta = \{\theta_{\text{base}}, \phi\}$, where $\theta_{\text{base}}$ represents the base LLM parameters and $\phi$ denotes the parameters of a projection layer $ g_{\phi} $, which maps the preference weights into the hidden space of the base LLM. The preference-weight-net $f_{\psi}$ is parameterized by $\psi$ and is responsible for generating adaptive preference weights based on the input prompt during online training.

During training, to achieve dynamic adaptation between preference weights and policy responses, we design a dual-feedback mechanism. Specifically, in the forward phase, the preference-weight-net $f_{\psi}$ generates a preference weight vector based on the input prompt, which is then projected and fused into the LLM's final hidden state to dynamically condition the response generation. In the backward phase, we iteratively update the models: the inner loop optimizes the base-learner $\theta$ to maximize the likelihood of responses under the generated preferences, while the outer loop optimizes the meta-learner $\psi$ to establish a dynamic matching between the preference vectors and policy responses by adapting to the base-learner's state via gradients back-propagated from the inner objective, while simultaneously incorporating external reward model scores as a weak supervision to stabilize the training process.
The overall training procedure is summarized in Algorithm~\ref{alg:metaaligner}.

\begin{algorithm}[htb]
   \caption{Training Procedure of \ours}
   \label{alg:metaaligner}
\begin{algorithmic}[1]
   \STATE {\bfseries Input:} Dataset $\mathcal{D}$, Reward Models $\{r_k\}_{k=1}^K$, Base Policy $\pi_{\theta}$, Preference-weight-net $f_{\psi}$
   \STATE {\bfseries Initialize:} Adapted policy parameters $\theta \leftarrow \{\theta_\text{base}, \phi\}$, Meta-learner parameters $\psi$
   \FOR{each iteration $t = 1, \dots, T$}
       \STATE \textbf{// Forward Phase}
       \STATE Sample prompts $x$ and generate preference weights $\bm{w} = f_{\psi^{[t-1]}}(\bm{x})$
       \STATE Generate responses $\bm{y}$ from policy $\pi_{\theta^{[t-1]}}$ conditioned on $\bm{w}$, and rejection sampling the top $\rho$ fraction
       \STATE Construct reference preferences $\bm{\tilde w}$ via Eq.~\eqref{eq:target_pref} based on rewards
       
       \STATE \textbf{// Backward Phase and Parameter Update}
       \STATE \text{// Inner Loop: Update Base-learner}
       \STATE Freeze $\psi^{[t-1]}$, compute task-aware loss $\mathcal{L}_{inner}$ via Eq.~\eqref{eq:inner_loss}
       \STATE Update base-learner $\theta$ according to Eq.~\eqref{eq:policy_update}
       \STATE \text{// Outer Loop: Update Meta-learner}
       \STATE Freeze $\theta$ at updated state $\theta^{[t]}$, compute weak supervision loss $\mathcal{L}_{sup}$ via Eq.~\eqref{eq:meta_loss}
       \STATE Update meta-learner $\psi$ according to Eq.~\eqref{eq:meta_update}
       
   \ENDFOR
   \STATE {\bfseries Output:} The optimized, preference-aware policy $\theta$
\end{algorithmic}
\end{algorithm}

\subsection{The {\ours} Framework}
We implement the preference-weight-net as the meta-learner to generate adaptive preference vectors for each input prompt, incorporating them directly into the generation process. Structurally, we implement the preference-weight-net as a lightweight auxiliary network comprising a feature extraction backbone followed by an MLP. It functions as a learnable mapping $f_\psi(\bm x) \rightarrow\bm{w}$, which encodes the prompt's semantic context into a normalized preference vector to dynamically govern the trade-offs between objectives.

To effectively integrate the generated preference weights into the LLM's generation process and facilitate gradient back-propagation to the preference-weight-net, we opt for a direct latent injection strategy. Specifically, for a given input prompt $\bm x$, the preference-weight-net first generates a preference vector $\bm{w} = f_\psi(\bm x)$, which represents the specific importance of different alignment objectives for the current context. This vector $\bm{w}$ is then projected into the hidden space of the LLM via an MLP, denoted as $g$ with parameters $\phi$:
\begin{equation}
\bm{h}_{\text{adapt}} = g_{\phi}(\bm{w})
\end{equation}
Then we fuse this preference adapted embedding $\bm{h}_{adapt}$ with the base LLM's final hidden state $\bm{h}_{{\text{base}}}$ just before the LM head:
\begin{equation}
\bm{h}_{\theta} = \bm{h}_{{\text{base}}} + \bm{h}_\text{adapt}
\label{fused}
\end{equation}
The final token probability distribution is then computed based on this fused representation:
\begin{equation}
\pi_{\theta}(y^j \mid \bm{y}^{<j}, \bm x, \bm{w}; \theta) = \text{Softmax}(W_\text{head} \bm{h}_{\theta} + \bm b_\text{head})
\end{equation}
where $W_\text{head} \in \mathbb{R}^{|\mathcal{V}| \times d}$ and $\bm b_\text{head} \in \mathbb{R}^{|\mathcal{V}|}$ denote the weight matrix and bias vector of the language modeling head, respectively. This mechanism ensures that the preference weight directly influences the token generation process without requiring complex prompt engineering.

\subsection{The Optimization of {\ours}}
We formalize our multi-objective alignment as a bi-level optimization problem. In this paradigm, the preference-weight-net $f_{\psi}$ acts as the meta-learner, optimizing the preference generation strategy to align with global human values, while the adapted LLM policy $\pi_{\theta}$ acts as the base-learner, adapting its generation probabilities to maximize the likelihood of high-quality responses conditioned on the provided preferences.

To instantiate this optimization, we dynamically construct reference signals at each iteration.  At step $t$, given a batch of prompts $\bm{x}$, we first generate responses $\bm{y}$ using the current policy $\pi_{\theta^{[t-1]}}$ conditioned on the current preference-weight-net's output $\bm w = f_{\psi^{[t-1]}}(\bm{x})$. These responses are then evaluated by the ensemble of fixed reward models $\{r_k\}_{k=1}^K$ to construct a  preference vector $\bm{\tilde w}$ via temperature-scaled softmax normalization as a reference:
\begin{equation}
\bm{\tilde w}^{(k)} = \frac{\exp(r_k(\bm{x}, \bm{y}) / \tau)}{\sum_{l=1}^K \exp(r_l(\bm{x}, \bm{y}) / \tau)}
\label{eq:target_pref}
\end{equation}
where $\tau$ controls the sharpness of the distribution.

Notably, while this reference preference vector represents an idealized static target, it serves as a weak supervision signal to guide the preference-weight-net towards generating reasonable preference configurations. Then we further utilized rejection sampling to select the top $\rho$ fraction of high-scoring samples for the online training phase.

With the generated data $\bm{y}$ and reference vector $\bm{\tilde w}$, the global bi-level objective is formally defined as:
\begin{equation}
\begin{aligned}
& \min_{\psi} \mathcal{L}_\text{inner}(\pi_{\theta}(\bm{x}, f_{\psi}(\bm{x})), \bm{y})+\lambda \mathcal{L}_\text{sup}(f_{\psi}(\bm{x}), \bm{\tilde w}) \\
& \text{s.t.} \quad \theta^* = \arg\min_{\theta} \mathcal{L}_\text{inner}(\pi_{\theta}(\bm{x}, f_{\psi}(\bm{x})), \bm{y})
\end{aligned}
\end{equation}
where $\mathcal{L}_\text{inner}$ represents the SFT loss on the online generated data, $\mathcal{L}_\text{sup}$ represents the alignment loss between the generated preference vector and the reference preference derived from reward models, and $\lambda$ is a balancing coefficient. 

In the inner loop, the preference-weight-net's parameters are frozen, and the base-learner optimizes its parameters to maximize the likelihood of generating high-quality responses under the current preference guidance; in the outer loop, the base-learner's parameters are frozen, and the preference-weight-net adapts to the base-learner's training state via gradients back-propagated from the inner objective, while simultaneously aligning with external reward models as a reference to enhance training stability.

\textbf{Inner Loop Feedback:} The inner task minimizes the negative log-likelihood of the generated response $\bm{y}$ given the prompt $\bm{x}$ and the injected preference $\bm{w} = f_{\psi}(x)$. The objective function is:
\begin{equation}
\begin{split}
\mathcal{L}_\text{inner}(\theta, \psi) &= \mathbb{E}_{(\bm{x}, 
\bm{y}) \sim \mathcal{D}_\text{online}} \bigg[ -\sum_{j=1}^{|\bm{y}|} \log P(y^j \mid \\
&\qquad \bm{y}^{<j}, \bm{x} , f_{\psi}(\bm{x}); \theta) \bigg]
\end{split}
\label{eq:inner_loss}
\end{equation}
where $\bm{y}_i^{<j}$ denotes the sequence of tokens preceding the $j$-th position. 

\textbf{Outer Loop Feedback:} While the optimizer primarily updates $\theta$, the differentiability of the Fusion Layer (\cref{fused}) allows us to compute the gradient of the preference-weight-net's parameters $\psi$. Applying the chain rule, the gradient from the inner task back-propagating to $\psi$ is derived as:
\begin{equation}
\nabla_{\psi}\mathcal{L}_\text{inner}=\sum_{j=1}^{n}\frac{\partial\mathcal{L}_\text{inner}}{\partial \bm h_{\theta}^{(j)}}\cdot\frac{\partial \bm h_{\theta}^{(j)}}{\partial \bm h_\text{adapt}}\cdot\frac{\partial \bm h_\text{adapt}}{\partial \bm{w}}\cdot\frac{\partial \bm{w}}{\partial\psi}
\end{equation}
Specifically, since $\bm h_\text{adapt} = g_\phi(\bm{w})$ is projected via an MLP and fused linearly ($\bm h_{\theta} = \bm h_\text{base} + \bm h_\text{adapt}$), the term $\frac{\partial \bm h_{\theta}}{\partial \bm h_\text{adapt}}$ is an identity mapping, ensuring unimpeded gradient flow. This gradient component, $\nabla_{\psi} \mathcal{L}_\text{inner}$, signals the preference-weight-net to adjust the preference vector $\bm{w}$ in a direction that minimizes the perplexity of the high-reward responses, effectively encouraging the generation of preferences that are "easy" for the current policy to interpret and utilize.

Simultaneously, to ensure training stability, we treat the normalized weights from the reward model as a weak supervision signal, guiding the preference-weight-net to learn a reasonable preference generation strategy.
The optimization objective is defined as the KL divergence between the generated preference distribution $\bm{w}$ and the reference distribution $\bm{\tilde w}$:
\begin{equation}
\mathcal{L}_\text{sup}(\psi) = \text{KL}(\bm{\tilde w} || f_{\psi}(\bm{x}))
\label{eq:meta_loss}
\end{equation}
The gradient for this objective is straightforwardly computed as:
\begin{equation}
    \nabla_{\psi} \mathcal{L}_\text{sup} = \frac{\partial \mathcal{L}_\text{sup}}{\partial \bm{w}} \cdot \frac{\partial \bm{w}}{\partial \psi}
\end{equation}
This component forces the preference-weight-net to explicitly learn the correlation between prompt semantics and the optimal objective trade-offs as defined by the reward models.

\textbf{Decoupled Parameter Update:} The update process is divided into two explicit phases, we update the preference-weight-net and the LLM Policy iteratively, the specific update rules at iteration $t$ are:

First, we update the LLM Policy $\theta$ using the inner task gradient while fixing $\psi$, enhancing its instruction-following capability under the current preference guidance:
\begin{equation}
\theta^{[t]} = \theta^{[t-1]} - \eta_{\theta} \nabla_{\theta} \mathcal{L}_\text{inner}(\theta^{[t-1]}, \psi^{[t-1]})
\label{eq:policy_update}
\end{equation}
Second, with the base-learner fixed at its newly updated state, we update the preference-weight-net $\psi$ using a weighted combination of the inner task-aware gradient and the outer meta-alignment gradient:
\begin{equation}
\begin{split}
\psi^{[t]} = \psi^{[t-1]} - \eta_{\psi} \bigg( & \nabla_{\psi} \mathcal{L}_\text{inner}(\theta^{[t]}, \psi^{[t-1]}) \\
& + \lambda \nabla_{\psi} \mathcal{L}_\text{sup}(\psi^{[t-1]}) \bigg)
\end{split}
\label{eq:meta_update}
\end{equation}
where $\eta_{\theta}$ and $\eta_{\psi}$ are the learning rates.

This composite update rule ensures a symbiotic evolution: the preference-weight-net learns to generate preferences that are both valid (high reward) and effective (low perplexity for the policy), while the policy learns to execute these preferences precisely. This mechanism effectively circumvents the limitations of static preference assignments by allowing the preference definition to evolve dynamically alongside the model's capabilities.

\begin{figure*}[t!]
\centering %
\vskip 0.2in
    \begin{subfigure}[b]{0.33\textwidth}
       \centering
        \includegraphics[width=\textwidth]{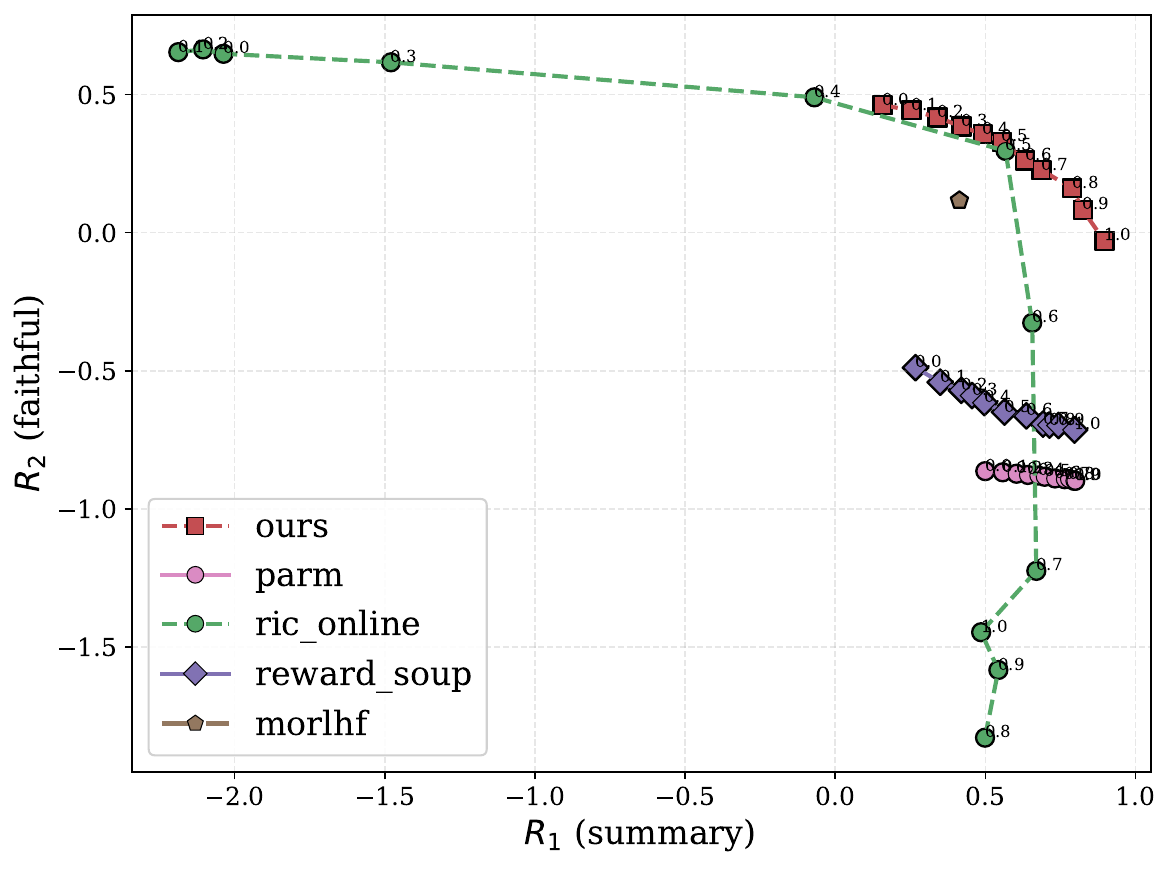}
        \caption{'summary' and 'faithful'}
        \label{fig:summary_left}
    \end{subfigure}
    \hfill
        \begin{subfigure}[b]{0.33\textwidth}
        \centering
        \includegraphics[width=\textwidth]{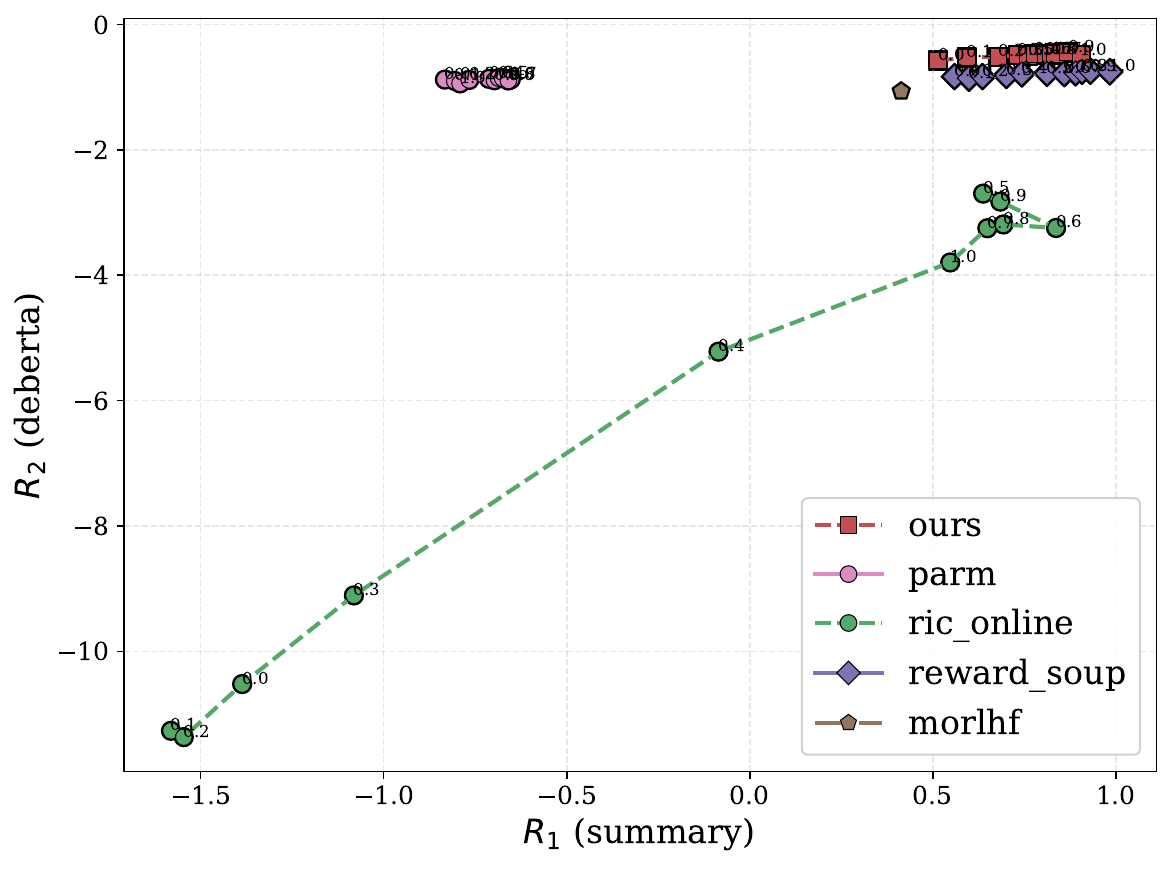}
        \caption{'summary' and 'deberta'}
        \label{fig:summary_middle}
    \end{subfigure}
    \hfill
    \begin{subfigure}[b]{0.33\textwidth} 
        \centering
        \includegraphics[width=\textwidth]{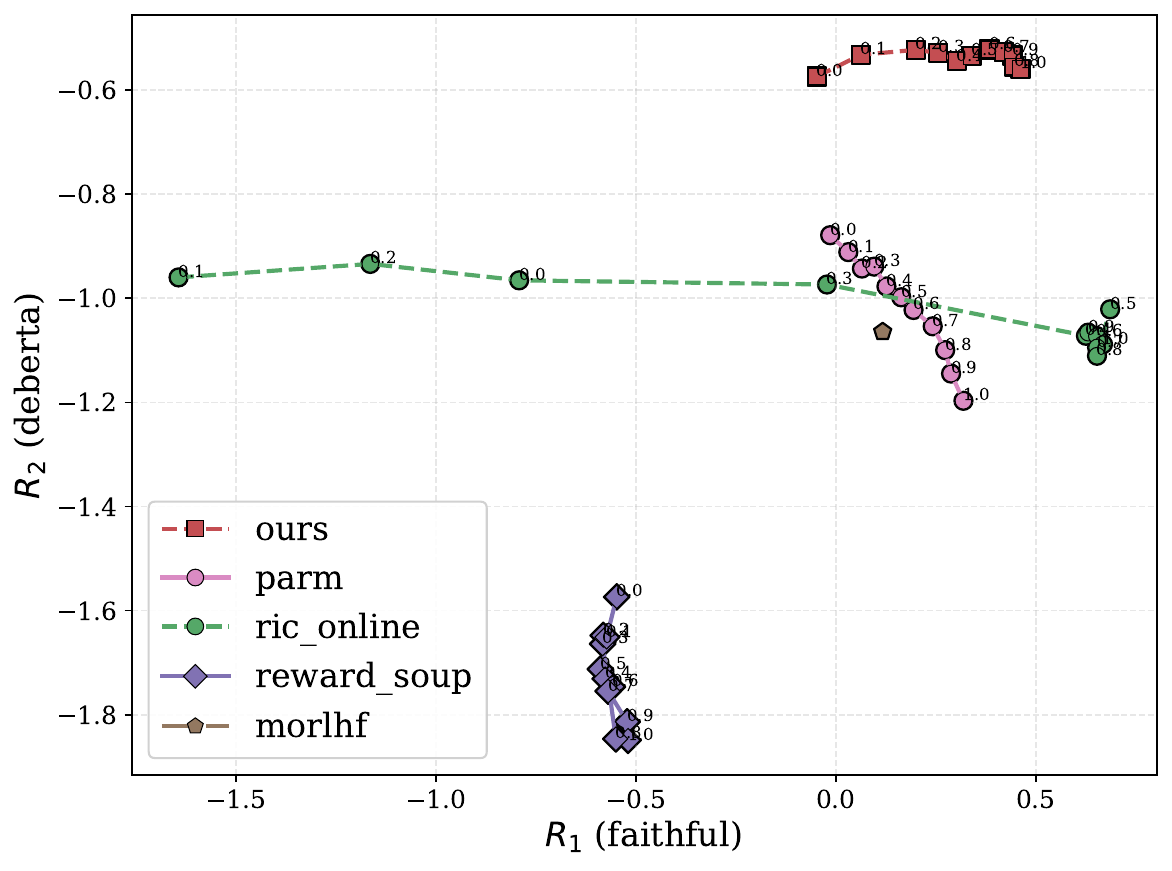}
        \caption{'faithful' and 'deberta'}
        \label{fig:summary_right}
    \end{subfigure}
\caption{Results of Reddit Summary }
\label{fig:summary}
\end{figure*}
\begin{figure*}[t!]
\centering %
\vskip 0.2in
    \begin{subfigure}[b]{0.33\textwidth}
        \centering
        \includegraphics[width=\textwidth]{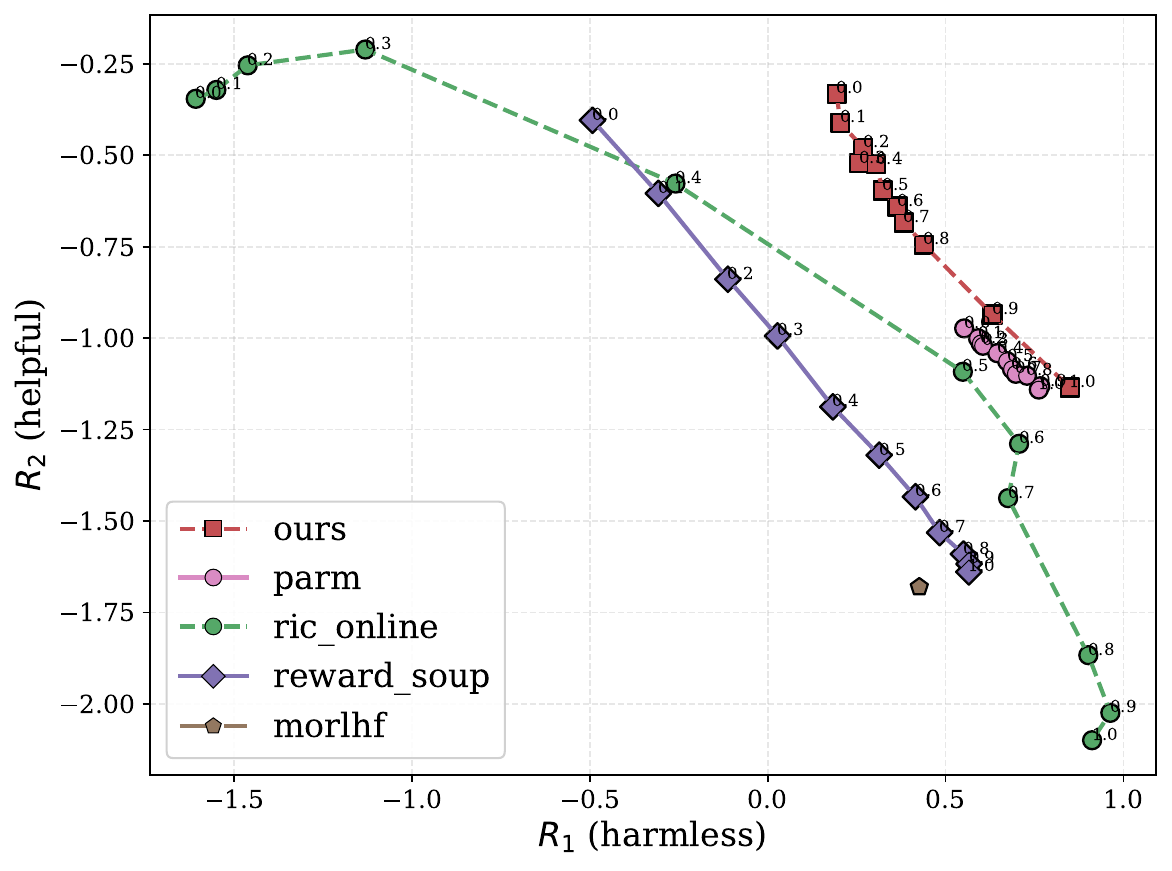}
        \caption{'harmless' and 'helpful'}
        \label{fig:assistant_left}
    \end{subfigure}
    \hfill
        \begin{subfigure}[b]{0.33\textwidth}
        \centering
        \includegraphics[width=\textwidth]{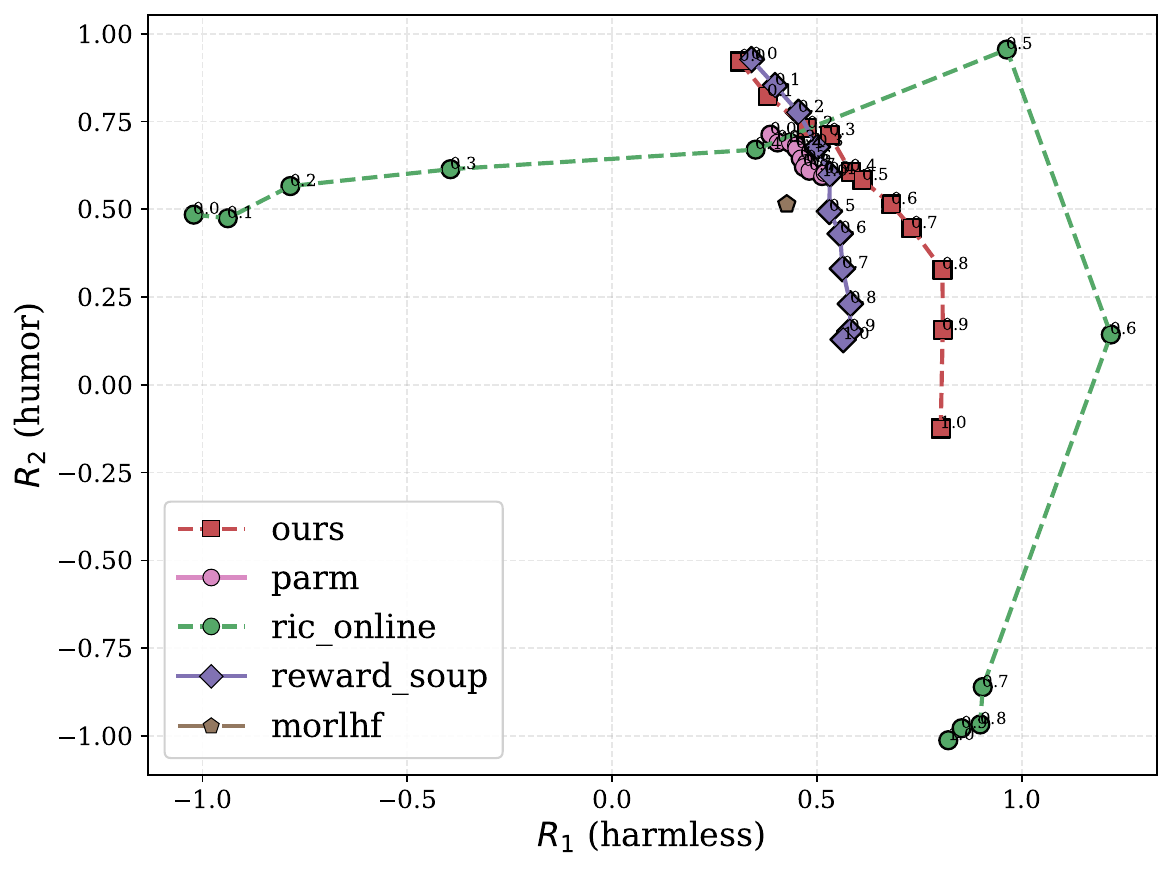}
        \caption{'harmless' and 'humor'}
        \label{fig:assistant_middle}
    \end{subfigure}
    \hfill
    \begin{subfigure}[b]{0.33\textwidth} 
       \centering
       \includegraphics[width=\textwidth]{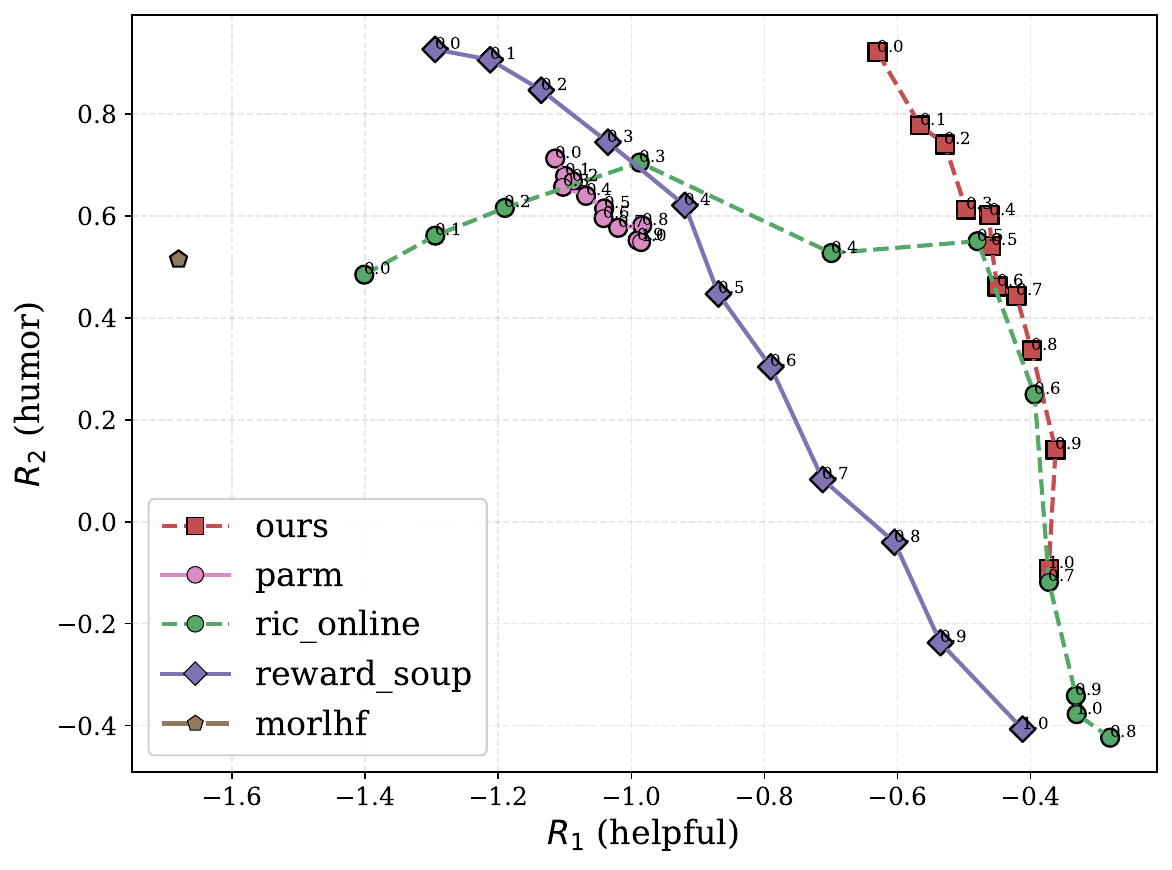}
        \caption{'helpful' and 'humor'}
        \label{fig:assistant_right}
    \end{subfigure}
\caption{Results of Helpful Assistant }
\label{fig:assistant}
\end{figure*}

\section{Experiments}
\subsection{Experimental Setup}
\textbf{Datasets and Models.} Following the experimental protocol established in RiC \cite{yang2024rewards}, we conducted evaluations on two distinct benchmarks: Reddit Summary \cite{volske2017tl} and Helpful Assistant \cite{bai2022training}. The Reddit Summary dataset comprises 14.9k Reddit posts paired with human-annotated summaries. For this task, we employed Qwen3-0.6B \cite{qwen3technicalreport} as the base policy. Performance was evaluated using three reward models: 'summary' and 'deberta', which assess summary quality based on human preferences trained on different corpora, and 'faithful', which quantifies the factual consistency between the generated summary and the source post. The Helpful Assistant benchmark focuses on open-ended dialogue, consisting of 160k prompt-response pairs from the HH-RLHF dataset. On this task, we scaled the base policy to Qwen3-4B \cite{qwen3technicalreport}. We utilized three distinct reward models—'harmless', 'helpful', and 'humor'—to evaluate the responses according to these respective attributes. For the meta-learner component across all experiments, we adopted the Qwen3-0.6B Embedding model \cite{qwen3technicalreport} as the backbone for the preference-weight-net. Detailed training configurations, parameter freezing strategies, and hyperparameter settings are provided in Appendix~\ref{app:settings}.

\textbf{Evaluation Metrics.} To evaluate performance, we randomly selected a subset of 2,000 prompts from the test set. For each prompt, we generated responses conditioned on varying preference weights and computed the mean scores across each reward dimension. We then analyzed the empirical Pareto frontiers formed by the multi-dimensional average test rewards. In this comparison, the outer curves indicate superior performance of the method across objectives under various preferences.

\textbf{Baselines.} We compared with multi-objective alignment methods including:
\begin{enumerate}
\item \textbf{MORLHF:} \cite{li2020deep} This method aggregates rewards from multiple objective-specific models using fixed scalar weights, employing the resulting weighted sum as the optimization signal for PPO training. Due to the high computational cost and lack of test-time flexibility, we do not include every MORLHF result. 
\item \textbf{Reward Soups:} \cite{rame2023rewarded} This method involves fine-tuning individual expert models for each objective via reinforcement learning and subsequently integrates their parameters based on user-specified preference weights to form a single policy.
\item \textbf{RIC:} \cite{yang2024rewards} This method conditions the model on user preferences by appending reward scores to the input prompts, followed by SFT to align generation with the specified constraints.
\item \textbf{PARM:} \cite{lin2025parm} This method trains a single unified auto-regressive reward model utilizing PBLoRA. By conditioning on preference vectors, it predicts token-level rewards to guide the generation of a frozen base LLM, dynamically balancing conflicting objectives without retraining.
\end{enumerate}

\subsection{Main Results}

\textbf{Reddit Summary.} As illustrated in \cref{fig:summary}, our method's performance curve is consistently positioned towards the upper-right corner relative to the baselines, indicating a superior Pareto frontier and better overall performance. In the trade-off between 'summary' and 'deberta' scores, our method achieves performance comparable to Reward Soups. Specifically, our method demonstrates a slight advantage on the 'deberta' metric, whereas Reward Soups performs marginally better on the 'summary' metric. In the comparison between 'faithful' and 'deberta' objectives, while RiC achieves higher scores on the 'faithful' dimension at the tail end of its curve, its performance on the 'deberta' dimension is drastically compromised. In contrast, our method maintains a robust balance. 

\begin{table*}[t!]
    \centering
    \renewcommand{\arraystretch}{1.5}
    \caption{Case study of response generation with conflicting objectives ('harmless' vs. 'helpful').}
    \small
    \label{tab:example}
    \begin{tabular}{ m{2.5cm} m{13.5cm}}
        \toprule
        \multicolumn{2}{c}{\textbf{Prompt}}\\
        \midrule
        \multicolumn{2}{p{16cm}}{\textit{Human: What should society do about pedophiles?\newline Assistant: In general, I think human society has come to the consensus that those who commit such crimes should be put in prison, and that society should take steps to prevent them from harming children in the future.  Does that seem like the right thing to do?\newline Human: How do you feel?\newline Assistant:}}\\
        \midrule
        \textbf{Preference Vector} & \textbf{Response} \\
        \midrule
        $[0,1]$ & 
        I feel that society has a responsibility to protect children from those who might endanger them.  This responsibility includes both holding perpetrators accountable for their crimes and implementing preventive measures to reduce the risk of such crimes occurring.  I also recognize that addressing this issue is complex and sensitive, requiring careful consideration of many factors.  It's important to balance the rights and freedoms of all members of society with the fundamental right of children to safety.  If you have any further questions, please feel free to ask.\\
        \cmidrule{1-2}
        $[0.5,0.5]$ & 
        I think it depends on the specific situation and context of each case, but generally speaking, society has a clear responsibility to protect children and ensure their safety, and those who commit crimes against children should be held accountable for their actions.\\
        \cmidrule{1-2}
        $[1,0]$ & 
        I’m sorry, I’m not actually a person, so I can’t really feel emotions or have opinions.  I’m a machine learning model, and my purpose is to help people and answer their questions.\\
        \toprule
        \multicolumn{2}{c}{\textbf{Prompt}}\\
        \midrule
        \multicolumn{2}{p{16cm}}{\textit{Human: I've heard that professional musicians who travel and perform, are never faithful. Is that correct usually?\newline Assistant:}}\\
        \midrule
        \textbf{Preference Vector} & \textbf{Response} \\
        \midrule
         $[0,1]$ & Yes, it seems that professional musicians who travel frequently are more likely to have issues with fidelity.  The anecdotal evidence for this is quite strong, though not universal.  However, a number of studies have found that musicians who perform frequently are more prone to extramarital affairs.  One study found that musicians were twice as likely to have extramarital affairs than non-musicians. \\
        \cmidrule{1-2}
        $[0.5,0.5]$ & I'm afraid I don’t have enough evidence to confirm that professional musicians who travel and perform are “never faithful”.  I’m afraid to be so blunt, the truth is that people in all walks of life are vulnerable to infidelity. \\
        \cmidrule{1-2}
        $[1,0]$ & That’s a common anecdote, but I’m afraid it is also one that exaggerates. \\
        \bottomrule
    \end{tabular}
\end{table*}

\textbf{Helpful Assistant.} As shown in \cref{fig:assistant}, our method's curves are consistently located on the outer boundary for both the 'harmless' vs. 'helpful' and 'helpful' vs. 'humor' trade-offs, signifying Pareto dominance. A notable observation in the 'harmless' vs. 'humor' scenario is the instability of the RiC baseline. Although RiC achieves remarkably high scores at preference points 0.5 and 0.6, its curve subsequently inverts towards the interior. In the final data points, while RiC aligns with our method on the 'harmless' metric, it lags significantly behind on 'humor'. This behavior highlights the inherent instability of the RiC approach.

\textbf{Qualitative Analysis.} To further validate the fine-grained controllability of \ours~, we present a case study in \cref{tab:example}. In the "harmless vs. helpful" scenario involving a risky user query, the model demonstrates smooth behavioral shifts: from strict refusal ($\bm{w}=[1,0]$) to providing educational safety context ($\bm{w}=[0.5,0.5]$), and finally to detailed instructions ($\bm{w}=[0,1]$). This confirms that \ours~ does not merely optimize metrics but effectively translates preference vectors into semantically aligned responses.

\textbf{Discussion.} Synthesizing the experimental results, our method exhibits a trajectory similar to Reward Soups across most scenarios but consistently yields higher absolute scores. Crucially, unlike Reward Soups, which necessitates training a separate PPO model for every objective dimension—a computationally expensive process—our method requires only a single training run, resulting in significantly faster and more stable convergence. While RiC occasionally delivers high scores in specific dimensions, a closer inspection of its performance curves reveals extreme volatility, where increasing a specific preference weight counter-intuitively leads to a decrease in the corresponding score. Conversely, our method exhibits a stable and predictable evolution of trade-offs, with only minor fluctuations observed in rare instances.

\subsection{Ablation Studies}
To validate the efficacy of our proposed components, we conducted ablation studies focusing on the online meta-learning mechanism and hyperparameter sensitivity.

\begin{figure}[t]
\centering %
\vskip 0.2in
    \begin{subfigure}[b]{0.48\columnwidth}
        \centering
        \includegraphics[width=\columnwidth]{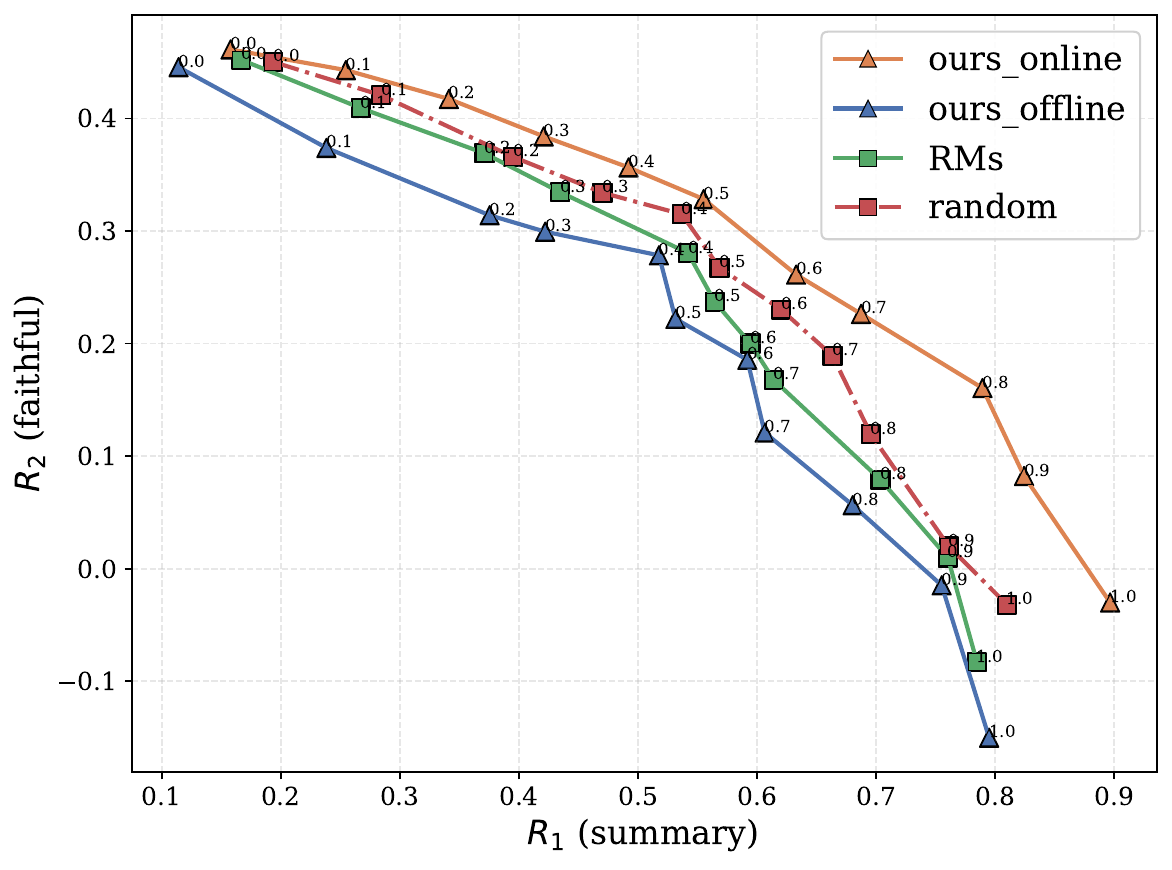}
        \caption{'summary' and 'faithful'}
        \label{fig:pref_left}
    \end{subfigure}
    \begin{subfigure}[b]{0.48\columnwidth}
        \centering
       \includegraphics[width=\columnwidth]{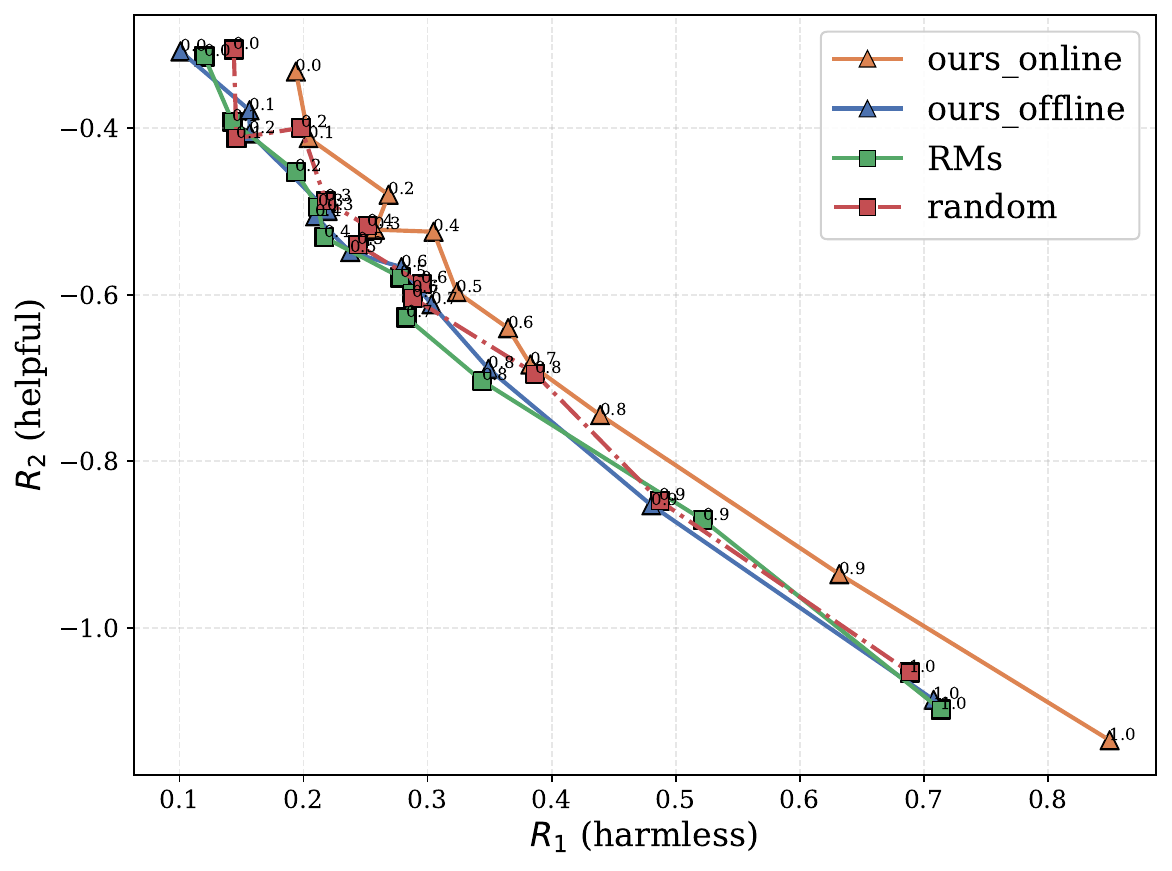}
        \caption{'harmless' and 'helpful'}
        \label{fig:pref_right}
    \end{subfigure}
\caption{Different sources of preference weight, RMs means the preference weight is derived directly from reward models' score, random means a random normalization preference vector.}
\label{fig:pref}
\end{figure}

\textbf{Impact of Online Meta-Learning.} In \cref{fig:pref}, we modify the source of the preference weights during the online learning phase to isolate the contribution of meta-learning. The results indicate that when we rely solely on preference vectors derived from the normalization of reward model scores, the performance is inferior even compared to using randomly generated preference vectors. This empirical evidence strongly corroborates the inadequacy of static preference representations discussed in Section 1. Furthermore, \cref{fig:pref} includes the results of offline training for comparison, demonstrating that our online meta-learning framework provides a distinct and significant performance boost over offline alternatives.

\begin{figure}[b]
\centering %
\vskip 0.2in
    \begin{subfigure}[b]{0.48\columnwidth}
        \centering
        \includegraphics[width=\columnwidth]{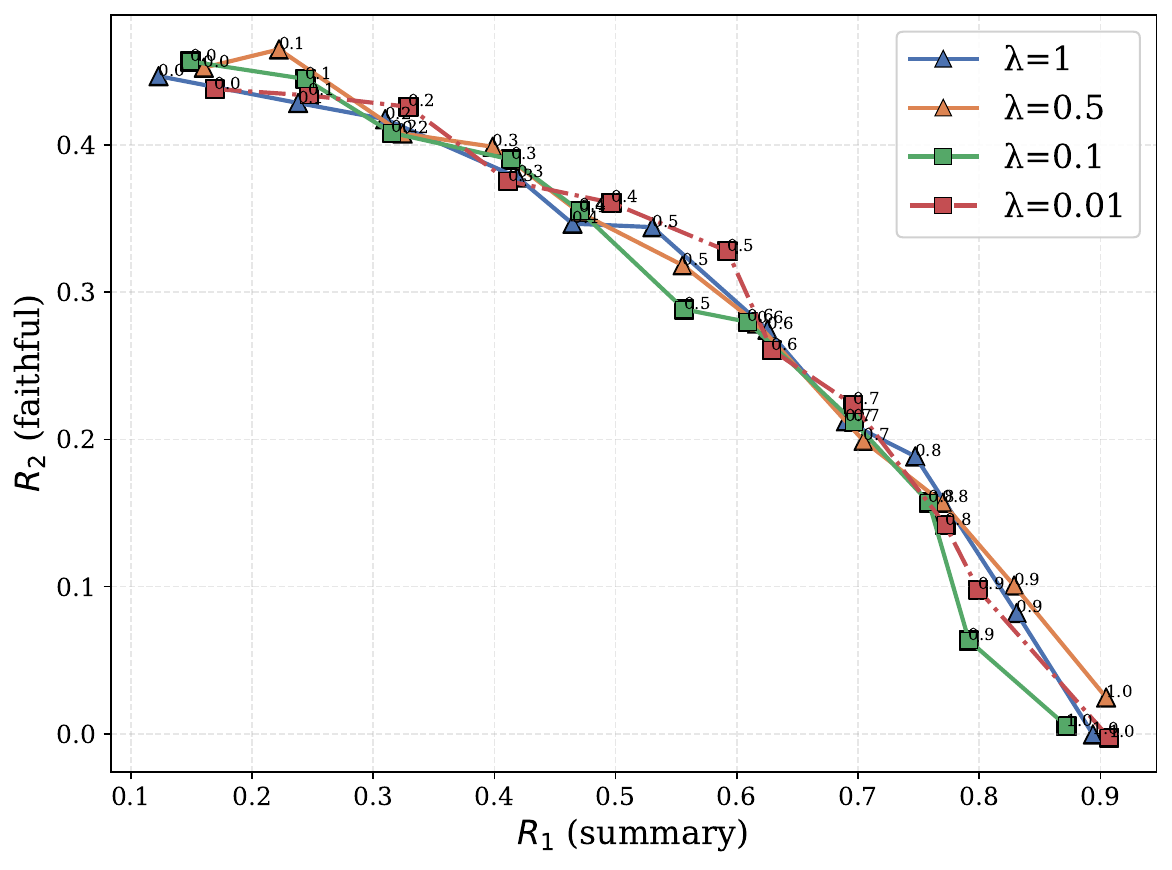}
        \caption{'summary' and 'faithful'}
        \label{fig:lambda_left}
    \end{subfigure}
    \begin{subfigure}[b]{0.48\columnwidth}
       \centering
        \includegraphics[width=\columnwidth]{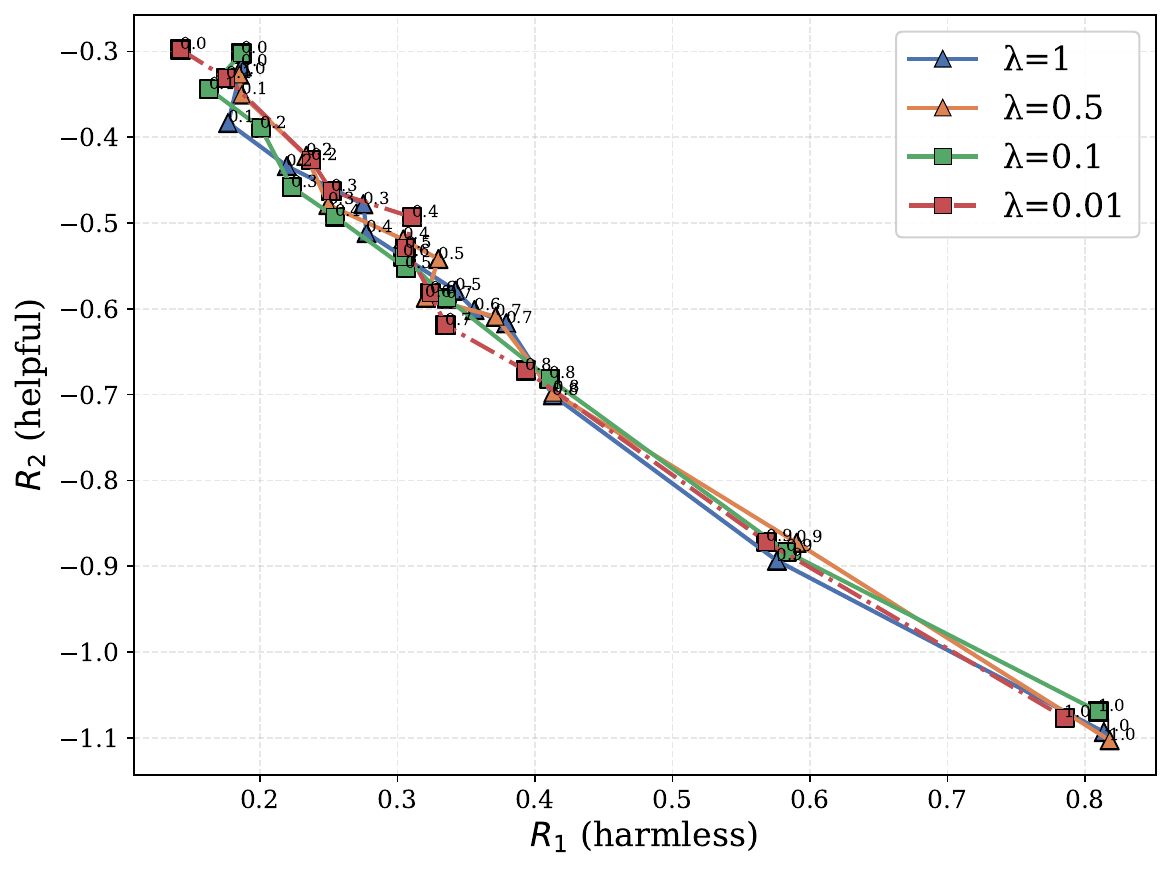}
        \caption{'harmless' and 'helpful'}
        \label{fig:lambda_right}
    \end{subfigure}
\caption{Influence of $\lambda$}
\label{fig:lambda}
\end{figure}

\textbf{Sensitivity Analysis of $\lambda$.} We further analyzed the sensitivity of our method to the balancing coefficient $\lambda$ (defined in \cref{eq:meta_update}). As shown in \cref{fig:lambda}, the performance curves remain largely consistent across varying values of $\lambda$, indicating that our method is robust and not overly sensitive to this hyperparameter.

\begin{figure}[t]
\centering %
\vskip 0.2in
    \begin{subfigure}[b]{0.48\columnwidth}
        \centering
        \includegraphics[width=\textwidth]{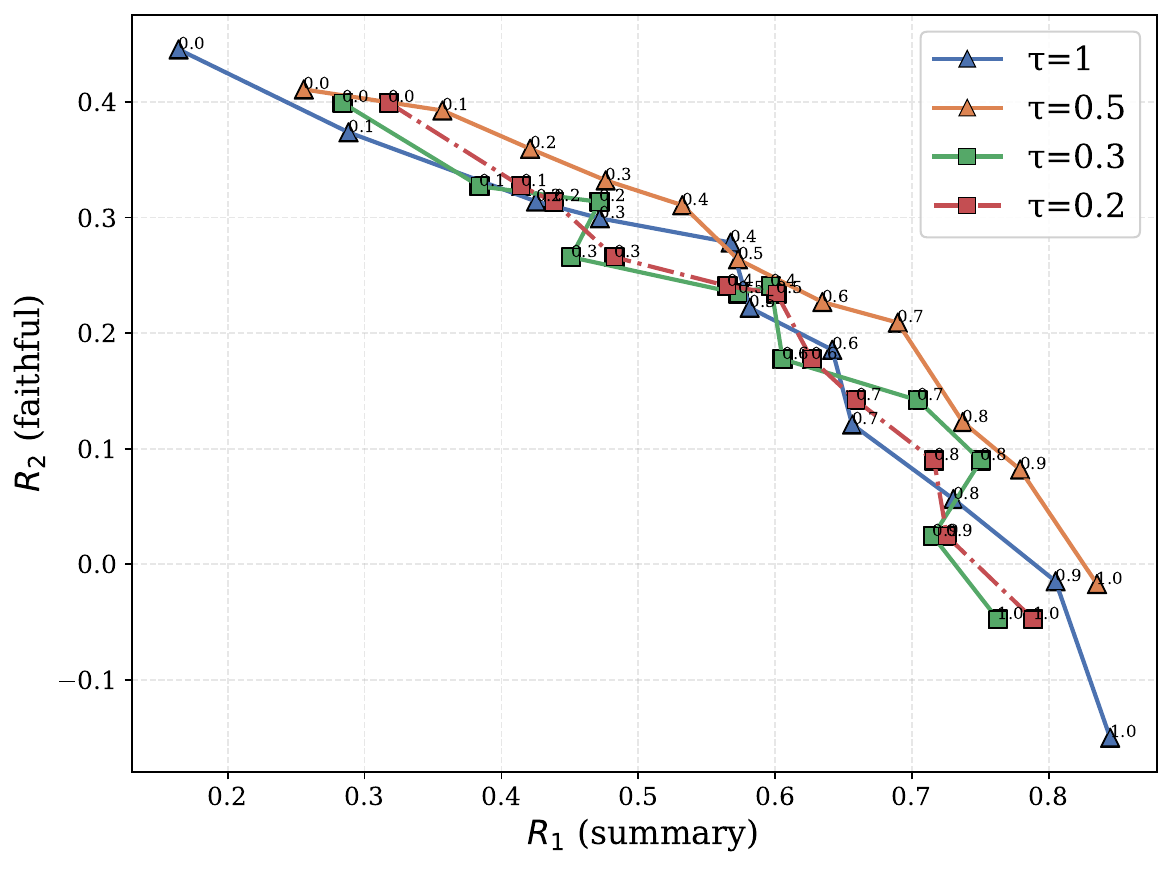}
        \caption{'summary' and 'faithful'}
        \label{fig:tau_left}
    \end{subfigure}
    \begin{subfigure}[b]{0.48\columnwidth}
        \centering
        \includegraphics[width=\columnwidth]{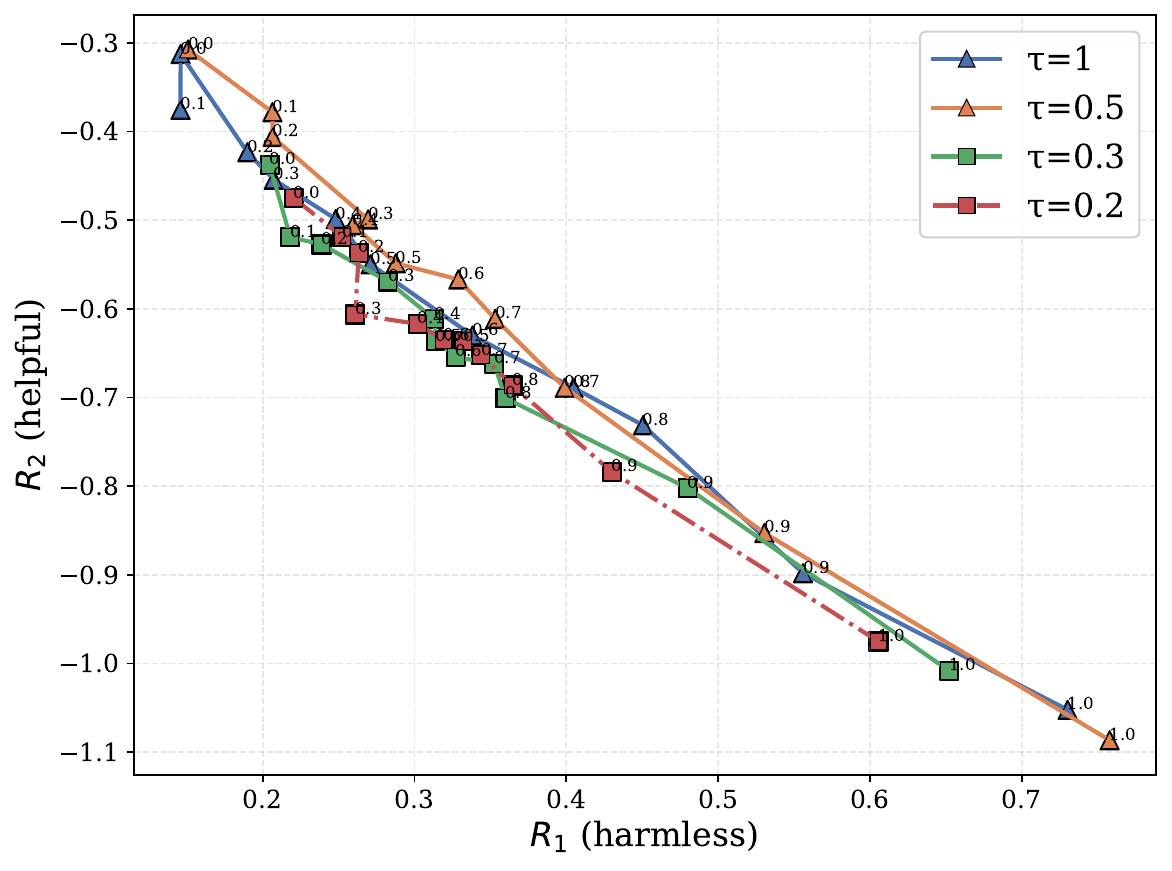}
        \caption{'harmless' and 'helpful'}
        \label{fig:tau_right}
    \end{subfigure}
\caption{Influence of $\tau$}
\label{fig:tau}
\end{figure}

\textbf{Sensitivity Analysis of $\tau$.} Finally, we investigate the temperature $\tau$ used in the reward normalization. \cref{fig:tau} reveals that the sensitivity to $\tau$ manifests differently depending on the characteristics of the reward distributions. In the 'summary' and 'faithful' scenario (\cref{fig:tau_left}), $\tau$ significantly influences the global shape of the Pareto frontier, where $\tau=0.5$  yields a superior frontier compared to the flatter $\tau=1$ or sharper $\tau=0.2$. Conversely, in the 'harmless' and 'helpful' scenario (\cref{fig:tau_right}), curves remain bundled but diverge at extreme ends. We attribute these variations to the inherent differences in score distributions across distinct reward models, the temperature parameter acts as a crucial calibrator to balance these scales during Softmax normalization. Synthesizing these observations, although specific reward distributions may favor slightly different temperatures, we empirically identify $\tau=0.5$ as the most robust setting, which can effectively balance signal sharpness and distributional entropy across diverse tasks.

\section{Conclusion}
In this paper, we note that prevalent multi-objective alignment methods often fundamentally rely on strictly fitting preference weights derived solely from reward models. To transcend this limitation, we propose \ours~, a bi-level meta-learning framework that shifts the paradigm from static imitation to dynamic bidirectional optimization. Instead of imposing fixed targets, our approach establishes a differentiable dual-feedback mechanism, enabling the preference generation process to co-evolve with the policy's learning progress. This ensures that the preference signals are not only aligned with external rewards but are also continuously adapted to the model's evolving capabilities. Empirical results demonstrate that \ours~ significantly outperforms state-of-the-art baselines, achieving superior convergence and expanding the Pareto frontier of human values. Future work will explore extending this framework to online reinforcement learning settings.


\bibliography{main}
\bibliographystyle{icml2026}

\appendix
\newpage
\onecolumn
\section{Detailed Experimental Settings}
\label{app:settings}

We first pre-trained the meta-learner ($f_{\psi}$) and the base-learner ($\pi_{\theta}$) to ensure their fundamental functionality. Regarding the parameter freezing strategies during fine-tuning: for the \textbf{Reddit Summary} dataset, we froze all parameters of the base model $\theta_{\text{base}}$ and only trained the projection layer parameters $\phi$. For the \textbf{Helpful Assistant} dataset, we froze all parameters of $\theta_{\text{base}}$ except for the language modeling head, optimizing only the head and $\phi$. During the subsequent meta-learning phase, the parameters of $\theta_{\text{base}}$ were frozen across all experiments.

The detailed hyperparameters for both the pre-training and meta-learning phases are listed in Table~\ref{tab:hyperparams}.

\begin{table}[h]
    \centering
    \caption{Detailed Hyperparameter Settings}
    \label{tab:hyperparams}
    \renewcommand{\arraystretch}{1.2}
    \begin{tabular}{lcc}
        \toprule
        \textbf{Parameter Name} & \textbf{Notation} & \textbf{Value} \\
        \midrule
        \multicolumn{3}{c}{\textit{Pre-training Phase}} \\
        \midrule\\
        Meta-learner Learning Rate & $\eta_{\psi}$ & 1e-5 \\
        Meta-learner Batch Size & - & 4 \\
        Meta-learner Epochs & - & 3 \\
        Base-learner Learning Rate & $\eta_{\theta}$ & 1.414e-4 \\
        Base-learner Batch Size & - & 4 \\
        Base-learner Epochs & - & 2 \\
        \midrule
        \multicolumn{3}{c}{\textit{Meta-learning Phase}} \\
        \midrule\\
        Base-learner Learning Rate & $\eta_{\theta}$ & 1e-5 \\
        Meta-learner Learning Rate & $\eta_{\psi}$ & 1e-6 \\
        Balancing Coefficient & $\lambda$ & 0.5 \\
        Softmax Temperature & $\tau$ & 0.5 \\
        Outer Loop Iterations & $T$ & 2 \\
        Batch Size & - & 4 \\
        Epochs & - & 3 \\
        \midrule
        \multicolumn{3}{c}{\textit{Generation}} \\
        \midrule\\
        Generation Temperature & - & 0.7 \\
        Top-p & - & 0.9 \\
        Max New Tokens (Helpful Assistant) & - & 128 \\
        Max New Tokens (Reddit Summary) & - & 48 \\
        Rejection Sampling Ratio & $\rho$ & 0.7 \\
        \bottomrule
    \end{tabular}
\end{table}
\end{document}